\documentclass[conference]{IEEEtran}
\usepackage{graphicx} 
\usepackage{amsmath} 
\usepackage{amssymb} 
\usepackage{placeins}
\usepackage{multirow}
\usepackage{array}
\usepackage{tabularx}
\usepackage{ragged2e}
\usepackage{algorithm}
\usepackage{algpseudocode}
\usepackage{threeparttable}
\usepackage{placeins}  
\usepackage{cite}
\usepackage{hyperref}
\usepackage{booktabs}
\usepackage{subcaption}
\usepackage{orcidlink}
\usepackage{todonotes}

\usepackage{tikz}
\usetikzlibrary{positioning, fit, backgrounds, arrows.meta, shapes.geometric, calc}

\title{Balancing Fidelity, Utility, and Privacy in Synthetic Cardiac MRI Generation: A Comparative Study}

\author{
    \IEEEauthorblockN{
        Madhura Edirisooriya\orcidlink{0009-0007-8166-3736}\IEEEauthorrefmark{1},  
        Dasuni Kawya\orcidlink{0009-0006-2921-4573}\IEEEauthorrefmark{1},  
        Ishan Kumarasinghe\orcidlink{0009-0004-3110-1290}\IEEEauthorrefmark{1}
        Isuri Devindi\orcidlink{0009-0005-6615-7937}\IEEEauthorrefmark{2}, \\
        Mary M. Maleckar\orcidlink{0009-0005-6615-7937}\IEEEauthorrefmark{4}, 
        Roshan Ragel\orcidlink{0000-0002-4511-2335}\IEEEauthorrefmark{1},
        Isuru Nawinne \orcidlink{0009-0001-4760-3533}\IEEEauthorrefmark{1},
        Vajira Thambawita\orcidlink{0000-0001-6026-0929}\IEEEauthorrefmark{3},  \\
    }
\IEEEauthorblockA{
  \IEEEauthorrefmark{1}University of Peradeniya, Sri Lanka \quad 
  \IEEEauthorrefmark{2}University of Maryland, College Park, USA \quad \\
  \IEEEauthorrefmark{4}Tulane University School of Medicine, USA; Simula Research Laboratory, Norway \quad
  \IEEEauthorrefmark{3}SimulaMet, Norway 
  }
}

\begin{document}

\maketitle

\begin{abstract}
Deep learning in cardiac MRI (CMR) is fundamentally constrained by both data scarcity and privacy regulations. This study systematically benchmarks three generative architectures: Denoising Diffusion Probabilistic Models (DDPM), Latent Diffusion Models (LDM), and Flow Matching (FM) for synthetic CMR generation. Utilizing a two-stage pipeline where anatomical masks condition image synthesis, we evaluate generated data across three critical axes: fidelity, utility, and privacy. Our results show that diffusion-based models, particularly DDPM, provide the most effective balance between downstream segmentation utility, image fidelity, and privacy preservation under limited-data conditions, while FM demonstrates promising privacy characteristics with slightly lower task-level performance. These findings quantify the trade-offs between cross-domain generalization and patient confidentiality, establishing a framework for safe and effective synthetic data augmentation in medical imaging.
\end{abstract}


\section{Introduction}

Medical imaging data plays a central role in the development of data-driven artificial intelligence (AI) systems for clinical decision support. In cardiac imaging, deep learning models have shown strong potential for tasks such as segmentation, functional assessment, and diagnosis. However, progress in this area is fundamentally constrained by the limited availability of high-quality annotated datasets~\cite{paper25}. Even when such annotated datasets are available, they are typically small, often comprising only tens to a few hundred patients, and frequently underrepresent rare cardiac pathologies~\cite{paper04, acdc}. Cardiac MRI (CMR) data are particularly difficult to access due to strict privacy regulations~\cite{paper10}, the high cost of expert annotation, and the logistical challenges of multi-center data sharing. As a result, many AI models are trained on small, institution-specific datasets that fail to capture the diversity of real-world clinical scenarios~\cite{paper04}. 

Beyond data scarcity, privacy concerns further restrict the direct use and dissemination of medical images due to regulatory frameworks such as GDPR and HIPAA ~\cite{voigt2017eu,hipaa1996} . Even when data are anonymized, medical images encode patient-specific anatomical characteristics that can lead to unintended identity leakage~\cite{paper02}. This creates a critical tension between the need for large, diverse datasets and the obligation to protect patient confidentiality. Addressing this challenge requires approaches that can expand training data while minimizing privacy risks.

Recent advances in deep generative modeling offer a promising solution to these limitations. By learning the underlying distribution of medical images, these models can synthesize realistic, anatomically plausible samples to augment scarce medical datasets. 
Diffusion-based models have emerged as the state-of-the-art for high-fidelity medical image generation~\cite{paper25}, while flow-matching models have recently gained attention as an efficient alternative with deterministic sampling~\cite{paper47}. However, their relative performance in terms of downstream task utility and privacy preservation remains insufficiently understood.

In this work, we investigate the application of diffusion and flow-matching models for synthetic cardiac MRI generation under constrained data settings. Rather than focusing solely on visual realism, we evaluate synthetic data both based on its practical usefulness for downstream segmentation tasks and its ability to mitigate privacy risks. By systematically comparing these generative approaches, we aim to determine how synthetic CMR data can be effectively leveraged to improve model generalization across datasets while maintaining clinically acceptable performance and patient privacy.  Our code is available at \url{https://github.com/vlbthambawita/SynCMRI}. The models from our research are hosted on Hugging Face at \url{https://huggingface.co/spaces/ishanthathsara/SynCMRIApp}.

\section{Related Work}

\textbf{Table}~\ref{tab:cross_dataset_comparison} summarizes prior work in synthetic CMR generation across three key dimensions: downstream utility, image fidelity, and privacy preservation. Most existing studies to date evaluate generative models primarily in terms of visual realism and segmentation performance, with diffusion-based and GAN-based approaches consistently demonstrating improvements in image quality and task utility.
\begin{table}[!htbp]
\caption{Comparison of Generative Approaches Across Utility, Fidelity, and Privacy}
\centering
\small
\begin{tabular}{|p{0.8cm}|p{2.7cm}|c|c|c|c|c|}
\hline
\textbf{Study} & \textbf{Model} & \textbf{U} & \textbf{F} & \textbf{P} & \textbf{CDU} & \textbf{LDU} \\
\hline
\cite{paper15} & Diffusion (SD-DM) & $\checkmark$ & $\checkmark$ & -- & $\checkmark$ & $\checkmark$ \\
\cite{paper11} & Diffusion (Attention) & -- & $\checkmark$ & -- & -- & $\checkmark$ \\
\cite{paper42} & Diffusion (DiffuSeg) & $\checkmark$ & $\checkmark$ & -- & $\checkmark$ & $\checkmark$ \\
\cite{paper34} & LDM vs GAN & $\checkmark$ & $\checkmark$ & -- & $\checkmark$ & $\checkmark$ \\
\cite{paper25} & Diffusion (SPADE) & $\checkmark$ & $\checkmark$ & -- & $\checkmark$ & $\checkmark$ \\
\cite{paper10} & Diffusion (MONAI) & -- & $\checkmark$ & -- & -- & -- \\
\textbf{Ours} & \textbf{DDPM + LDM + FM} & $\checkmark$ & $\checkmark$ & $\checkmark$ & $\checkmark$ & $\checkmark$ \\
\hline
\end{tabular}
\label{tab:cross_dataset_comparison}
\begin{tablenotes}
\footnotesize
\item U: Utility, F: Fidelity, P: Privacy, CDU: Cross-Dataset Utility Ckeck, LDU: Limited Data Used.
\end{tablenotes}
\end{table}

\begin{figure*}[!htbp]
\centering
\includegraphics[width=\textwidth]{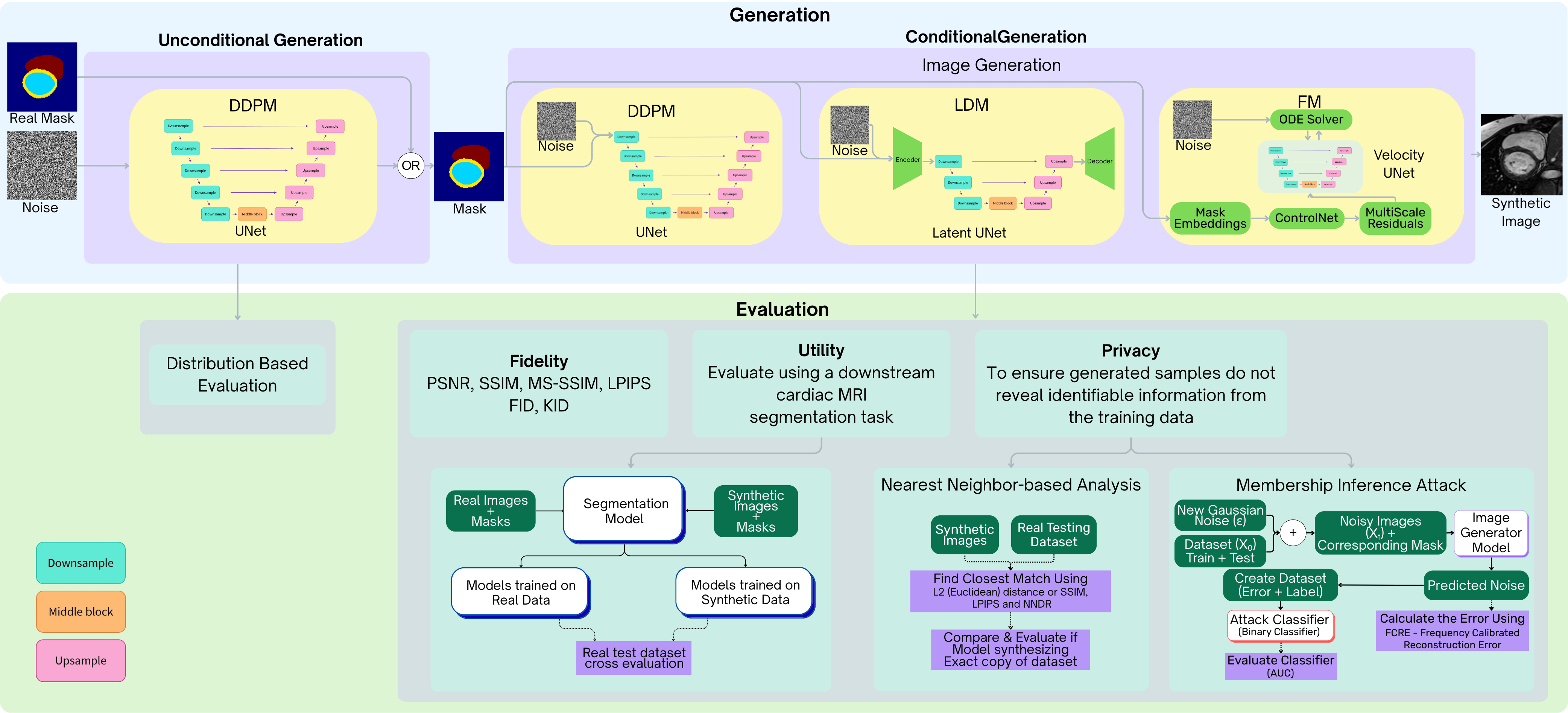}
\caption{Overview of the proposed synthetic cardiac MRI generation and evaluation framework. The pipeline consists of two main stages: (1) segmentation mask generation using a diffusion-based model, followed by (2) mask-conditioned image synthesis using DDPM, LDM, and Flow Matching. The generated images are evaluated across three dimensions: fidelity (PSNR, SSIM, MS-SSIM, LPIPS, FID, KID), downstream utility via cross-dataset cardiac MRI segmentation, and privacy through nearest-neighbor analysis and membership inference attacks.}
\label{fig:methodology}
\end{figure*}
However, privacy preservation is less systematically addressed, and only a limited subset of work explicitly incorporates privacy-aware mechanisms or formal privacy evaluation strategies. Furthermore, while diffusion and GAN models have been extensively studied in isolation, direct and controlled comparisons between diffusion-based and flow-matching paradigms remain scarce, particularly under limited-data and cross-dataset generalization settings. This highlights a critical gap in the literature: the absence of a unified evaluation framework in CMR generation that simultaneously assesses utility, fidelity, and privacy trade-offs across generative paradigms and cross-dataset conditions. 

\section{Methodology}
This research systematically compares diffusion-based and flow-matching generative models to determine which approach most effectively improves segmentation accuracy (utility), enhances cross-dataset generalization (fidelity), and preserves patient privacy in synthetic CMR generation (\textbf{Fig.}~\ref{fig:methodology}).

\subsection{Mask Generation}
The segmentation mask generator is based on a Denoising Diffusion Probabilistic Model (DDPM), which formulates mask generation as a stochastic Markov process that gradually transforms structured segmentation masks into Gaussian noise and learns to reverse this process.
\subsection{Image Generation}
After generating anatomically consistent segmentation masks in the preceding stage, corresponding CMR images are synthesized using mask-conditioned deep generative models. Conditioning on segmentation masks allows explicit enforcement of anatomical structure during image synthesis, ensuring consistency between generated textures and underlying cardiac morphology.

In this work, three classes of generative approaches are investigated: DDPM, Latent Diffusion Models (LDM), and Flow Matching (FM). While all three methods transform samples from a simple base distribution into realistic medical images, they differ fundamentally in how generation is formulated and conditioned on anatomical priors.

\subsubsection{Denoising Diffusion Probabilistic Model}
\begin{figure}[!htbp]
\centering
\includegraphics[width=\columnwidth]{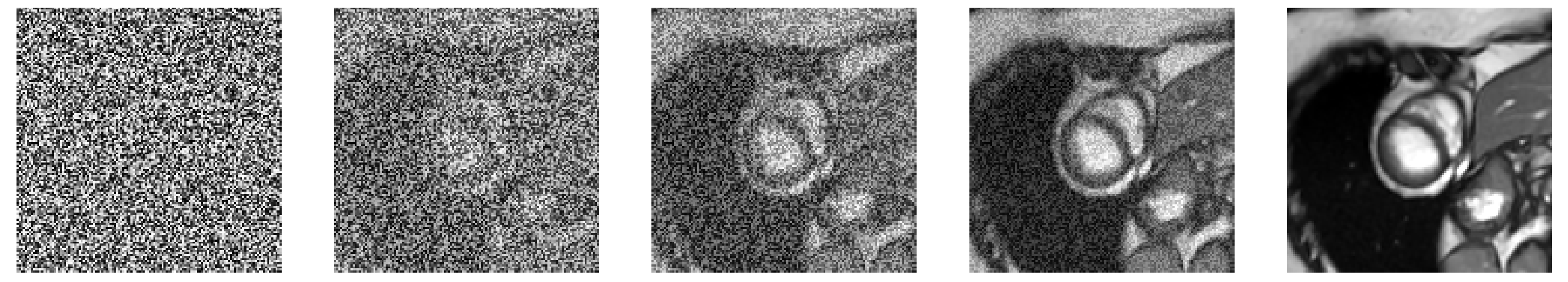}
\caption{Denoising Process. An image is shown each 200 timesteps}
\label{fig:denoisingProcess}
\end{figure}

DDPMs model image generation ~\cite{paper31} as a stochastic Markov process that gradually corrupts data with Gaussian noise and learns to reverse this corruption (\textbf{Fig.}~\ref{fig:denoisingProcess}). Given a real image \(x_0 \sim p_{\text{data}}\), the forward diffusion process is defined as,
\[
q(x_t \mid x_{t-1}) = \mathcal{N}\left(x_t; \sqrt{1 - \beta_t} x_{t-1}, \beta_t I\right), \quad t = 1, \dots, T,
\]

where \(\beta_t\) is a predefined noise schedule. This process admits a closed-form expression,
\[
x_t = \sqrt{\bar{\alpha}_t} x_0 + \sqrt{1 - \bar{\alpha}_t} \epsilon, \quad \epsilon \sim \mathcal{N}(0, I),
\]

with
\[
\bar{\alpha}_t = \prod_{s=1}^t (1 - \beta_s).
\]

The reverse generative process is parameterized by a neural network $\epsilon_\theta$, trained to predict the injected noise while being conditioned on the segmentation mask \(c\),
\[
p_\theta(x_{t-1} \mid x_t, c) = \mathcal{N}\left(x_{t-1}; \mu_\theta(x_t, t, c), \sigma_t^2 I\right).
\]

Training minimizes the denoising objective,
\[
\mathcal{L}_{\text{DDPM}} = \mathbb{E}_{x_0, \epsilon, t} \left[ \left\| \epsilon - \epsilon_\theta(x_t, t, c) \right\|_2^2 \right].
\]

By injecting the segmentation mask as an additional conditioning channel at each denoising step, the model is guided to reconstruct images that align anatomically with the generated masks.

\subsubsection{Latent Diffusion Model}
LDMs~\cite{paper23} address the computational limitations of pixel-space diffusion by performing the diffusion process in a compressed latent space. An autoencoder is first trained to encode an image \(x\) into a latent representation 
\(z = E(x)\) and reconstruct it via a decoder \(\mathcal{D}(z)\), preserving semantic and anatomical information while reducing spatial dimensionality.\\

The diffusion process is then applied in latent space,
\[
z_t = \sqrt{\bar{\alpha}_t} z_0 + \sqrt{1 - \bar{\alpha}_t} \epsilon
\]

The denoising network $\epsilon_\theta$ is trained to predict noise in the latent domain while conditioned on the segmentation mask \(c\),
\[
\mathcal{L}_{\text{LDM}} = \mathbb{E}_{z_0, \epsilon, t} \left[ \left\| \epsilon - \epsilon_\theta(z_t, t, c) \right\|_2^2 \right]
\]

After the denoising process, the generated latent representation is decoded back into image space using the decoder,
\[
\hat{x}_0 = D(z_0)
\]


This formulation of latent-space significantly reduces memory usage and inference time while maintaining high anatomical fidelity. Conditioning in the latent domain allows efficient integration of segmentation guidance, making LDMs particularly well suited for high-resolution cardiac MRI synthesis.\\

\subsubsection{Flow Matching}

FM models image generation as a deterministic, continuous-time transformation from a base distribution \( p_0 \) to the data distribution \( p_{\text{data}} \). It learns a time-dependent vector field \( v_\theta(x,t) \) governing the ODE
\begin{equation}
\frac{d x_t}{d t} = v_\theta(x_t, t),
\end{equation}
which is numerically integrated from an initial noise sample \( x_0 \sim p_0 \) to generate synthetic images.

We adopt the Optimal Transport Flow Matching (OT-FM) approach~\cite{paper45}, which defines probability paths between \( p_0 \) and \( p_{\text{data}} \) using optimal transport, resulting in efficient sampling and high-fidelity image generation.

\subsection{Evaluation methods}
\subsubsection{Synthetic Mask Evaluation}
Generated masks were evaluated using distribution-based shape analysis by comparing relative area and geometric properties of cardiac structures between real and synthetic samples, ensuring anatomical plausibility without relying on paired ground truth~\cite{paper11}.

\subsubsection{Synthetic MRI Fidelity Evaluation}
Fidelity evaluation assesses how closely synthetic CMR images resemble real images in terms of visual realism, structural integrity, and statistical distribution. In the context of medical image synthesis, high fidelity implies that generated images preserve clinically meaningful anatomical structures (e.g., ventricles and myocardium), realistic texture characteristics (e.g., contrast), and global distributional properties observed in real datasets.

In this work, fidelity is evaluated using both pixel-level and distribution-level metrics. Pixel-level similarity is quantified using SSIM, MS-SSIM and PSNR. Distribution-level realism is assessed using FID and KID, while perceptual similarity is measured using LPIPS. Together, these metrics provide a comprehensive assessment of structural, perceptual, and statistical fidelity of the generated CMR images.





\subsubsection{Utility Evaluation via Cardiac MRI Segmentation}

The utility of synthetic images is evaluated using a downstream cardiac MRI segmentation task, where segmentation performance on real data serves as an evaluation criterion for the practical usefulness of the generated datasets.

All experiments use a DynUNet segmentation architecture trained with a combined Dice and Cross-Entropy loss (DiceCE), both implemented via the MONAI library~\cite{monai} to ensure fair and consistent evaluation. All models are evaluated on both M\&Ms~\cite{paper21} and ACDC~\cite{acdc} test datasets to ensure fair and consistent comparison.

As reference baselines, segmentation models are trained using only real images. A model trained on the M\&Ms training set is evaluated on the M\&Ms test set for in-domain performance and on the ACDC test set for cross-dataset generalization. Similarly, a model trained on the ACDC training set is evaluated on both the ACDC and M\&Ms test sets. These experiments establish baseline segmentation performance using real annotated data and quantify the domain gap between the two datasets. To evaluate synthetic data utility, separate segmentation models are trained on synthetic images generated by DDPM, LDM, and FM, with one dedicated model trained per generative method.





\subsubsection{Synthetic CMR Privacy Evaluation}
To ensure patient privacy, synthetic medical image generation must not reveal identifiable information from the original training data. This risk is evaluated by testing whether the image generative models have simply memorized the real patient data. Two distinct evaluation methods are employed to ensure a fair comparison across all models.



\paragraph{Nearest Neighbor-based Privacy Analysis}
This method checks if the generated images are direct copies of the real patients used during training. Every synthetic image is compared against the training dataset to find its closest match~\cite{paper33}.

Similarity is measured using pixel difference (L2 distance) and visual perception (LPIPS). If a synthetic image is too similar to a real training image (very low distance), it indicates a privacy breach. The Nearest Neighbor Distance Ratio (NNDR) is also calculated  to confirm that the model has learned to generate unique variations rather than copying specific individuals.

\paragraph{Membership Inference Attack (MIA)}
This method tests if the model behaves differently when processing training images versus unseen test images~\cite{paper34}. If a model has memorized the training data, it will be able to reconstruct or denoise training images more accurately than unseen images. This performance gap is measured using the AUC score. A score of 0.5 is ideal, as it means the model treats training and new data exactly the same, whereas a high score indicates the model is leaking information about the training set.

Both methods are used to capture the full spectrum of privacy risks. While the Nearest Neighbor analysis detects visual replicas, the Membership Inference Attack detects statistical memorization. Together, they confirm that the model creates truly unique synthetic images.


\section{Experiments}
\subsection{Datasets}
We conduct experiments on two publicly available cardiac MRI benchmarks, ACDC~\cite{acdc} and M\&Ms~\cite{paper21}, which differ in size, vendor diversity, and acquisition settings. ACDC provides single-center, protocol-specific data for evaluating performance on high-quality unseen cases, while M\&Ms enables assessment of cross-vendor generalization across multi-center acquisitions.

\begin{table}[!htbp]
\caption{Comparison of ACDC and M\&Ms Cardiac MRI Datasets}
\centering
\scriptsize %
\renewcommand{\arraystretch}{1.1}
\setlength{\tabcolsep}{3pt}
\begin{tabularx}{\linewidth}{|p{1.4cm}|X|X|}

\hline
\textbf{Feature} & \textbf{ACDC} & \textbf{M\&Ms} \\
\hline
Total Subjects & 150 & 375 \\
\hline
Centres & Single Centre & Six Centres \\
\hline
Vendors & Single (Siemens, protocol-specific) & Four (Siemens, Philips, GE, Canon) \\
\hline
Countries & 1 (France) & 3 (Spain, Canada, Germany) \\
\hline
Target Labels & LV, RV, MYO (ED \& ES phases) & LV, RV, MYO (All the phrames) \\
\hline
Standard Utility & 
\begin{tabular}[c]{@{}l@{}}To test model performance on\\ high-quality unseen data\end{tabular}
&
\begin{tabular}[c]{@{}l@{}}Benchmarking cross-vendor\\ generalisation\end{tabular} \\
\hline
\end{tabularx}
\label{tab:acdc_mms_comparison}
\end{table}

\subsection{Data Preparation}
A two-stage preprocessing pipeline was employed to ensure spatial consistency, noise robustness, and stable model training. In the first stage, raw 4D NIfTI cardiac MRI volumes were processed frame-wise. N4 bias field correction was applied to mitigate intensity inhomogeneity caused by scanner artifacts~\cite{paper11}. Volumes were then resampled to a standardized in-plane resolution (1.25 × 1.25 mm) ~\cite{paper09} while retaining subject-specific through-plane spacing, ensuring cross-subject spatial consistency. Processed frames were stacked and cached, and only valid slices containing cardiac structures and sufficient intensity information were retained.

In the second stage, cached volumes were converted into 2D slices for training. A heart-centered region of interest (ROI) was extracted using the segmentation masks to focus learning on relevant anatomy and reduce background variability. Finally, percentile-based intensity normalization (1st–99th percentile) was applied and scaled to the range [-1, 1] to minimize the effect of outliers and improve optimization stability~\cite{paper15}.





\subsection{Implementation Details}
\subsubsection{Mask Generation}

In the first stage of the proposed pipeline, cardiac segmentation masks are generated using a diffusion-based generative model~\cite{monai}. We employ a 2D U-Net–based diffusion architecture with residual blocks and self-attention at deeper feature levels to capture both local anatomical details and global structural dependencies. The input segmentation masks are represented using one-hot encoding across four classes (background, left ventricle, myocardium, and right ventricle) and scaled to the range [-1,1]. 
The forward diffusion process follows a standard DDPM formulation with 1000 discrete timesteps ($t \in [0, 999]$), implemented using a linear Gaussian noise schedule. During training, for each batch, a timestep is randomly sampled uniformly from the 1000-step schedule, and Gaussian noise is added to the clean mask accordingly. The network is trained to predict the injected noise using a MSE objective. Optimization is performed using the Adam optimizer with a learning rate of $1 \times 10^{-4}$ over 400 epochs.
During inference, segmentation masks are generated by iterative reverse diffusion over the full 1000 timesteps, starting from Gaussian noise and progressively denoising to obtain anatomically plausible multi-class segmentation masks.



\subsubsection{Image Generation}
In the second stage of the pipeline, synthetic CMR images are generated using mask-conditioned generative models. A comprehensive description of the architectural design choices and optimization hyperparameters adopted for each generative framework is presented below.

\paragraph{Denoising Diffusion Probabilistic Model (DDPM)} is implemented using a UNet backbone. The model is designed to synthesize single-channel cardiac MRI images conditioned on multi-class segmentation masks.

The denoising network is based on a 2D UNet architecture implemented via the \textit{Diffusers} framework. The input consists of five channels, formed by concatenating the noisy image (1 channel) with the corresponding segmentation mask (4 channels). The network outputs a single-channel noise prediction.

The UNet architecture consists of six resolution levels with two layers per block and progressive feature widths of [128, 128, 256, 256, 512, 512]. Downsampling is performed via convolutional blocks, with a self-attention module introduced at the fifth resolution level to capture global anatomical context. The decoder mirrors the encoder and includes one attention-augmented upsampling block. All layers employ \textit{SiLU} activations. The forward diffusion process uses 1000 timesteps with a predefined Gaussian noise schedule. Segmentation conditioning is applied by concatenating mask channels to the noisy input at each timestep.

Training is performed using the \textit{AdamW} optimizer with an initial learning rate of $1 \times 10^{-4}$ and a cosine learning rate scheduler. The model is trained for 400 epochs with a batch size of 8 using multi-GPU data parallelism.

Image synthesis is conducted through iterative reverse diffusion over 1000 steps, enabling segmentation-guided sampling via a custom diffusion pipeline.

\paragraph{Latent Diffusion Model (LDM)}
is implemented to improve computational efficiency while preserving anatomical fidelity by performing diffusion in a learned \textbf{latent space} rather than directly in pixel space. The framework consists of two main components: a \textit{VQ-VAE autoencoder} for latent representation learning and a conditional UNet-based diffusion model operating on these latents.

A Vector-Quantized Variational Autoencoder (VQ-VAE) encodes 128×128 cardiac MRI images into 16×16 latent tensors with 4 channels across three resolution levels using progressive widths [64, 128, 256], with two residual blocks per level and \textit{Group Normalization} + \textit{SiLU} activations. The decoder reconstructs images using reconstruction and commitment losses. After training, the VQ-VAE is frozen for latent diffusion.

Diffusion operates in latent space using a conditional UNet with channel widths [128, 256, 512] and two downsampling stages. Segmentation masks are concatenated with noisy latents and integrated via SPADE-based conditioning, while sinusoidal timestep embeddings provide temporal encoding. Self-attention is applied at deeper resolutions.

The model follows the DDPM formulation with 1000 timesteps and a linear noise schedule. The UNet predicts injected noise using an MSE loss, optimized with \textit{AdamW} (learning rate $2 \times 10^{-4}$) and EMA stabilization.

During inference, latents are denoised over 1000 steps and decoded through the frozen VQ-VAE to produce anatomically consistent cardiac MR images.

\paragraph{Flow Matching}:
The model employs a 2D U-Net based velocity network augmented with ControlNet conditioning for segmentation-guided generation, operating on single-channel cardiac MRI images to predict the velocity field used during ODE-based sampling.

The U-Net backbone adopts a four-level encoder--decoder design with feature widths $[32, 64, 128, 256]$, two residual blocks per level, Group Normalization, and \textit{SiLU} activations. Self-attention is applied only at the deepest level using Spatial Transformer blocks for efficient global feature modeling.

Segmentation conditioning is implemented via a ControlNet-style residual guidance mechanism. A lightweight convolutional embedding of the segmentation mask is processed through a ControlNet branch mirroring the U-Net encoder and bottleneck, producing multi-scale additive residual features injected into the corresponding U-Net blocks. The ControlNet is initialized using partial weight sharing from the UNet backbone to improve training stability.

The continuous time variable \( t \in [0,1] \) is discretized into \( T = 1000 \) timesteps to reuse standard diffusion-style timestep embeddings within the U-Net, enabling seamless integration without architectural modifications.

Models are trained using the \textit{Adam} optimizer with a learning rate of $2.5 \times 10^{-5}$ for 100 epochs on GPU. During inference, synthetic images are generated by ODE-based integration of the learned velocity field from Gaussian noise, with segmentation masks provided to the ControlNet branch at each solver step.
 
\subsection{Computing Resources}
We trained and evaluated all models on a GPU server with three NVIDIA RTX 6000 Ada Generation GPUs. Each GPU provides approximately 48~GB of memory. The server used NVIDIA driver 580.82.07 and CUDA 13.0.

\section{Results \& Discussion}

\begin{figure}
    \centering
    \includegraphics[width=1\linewidth]{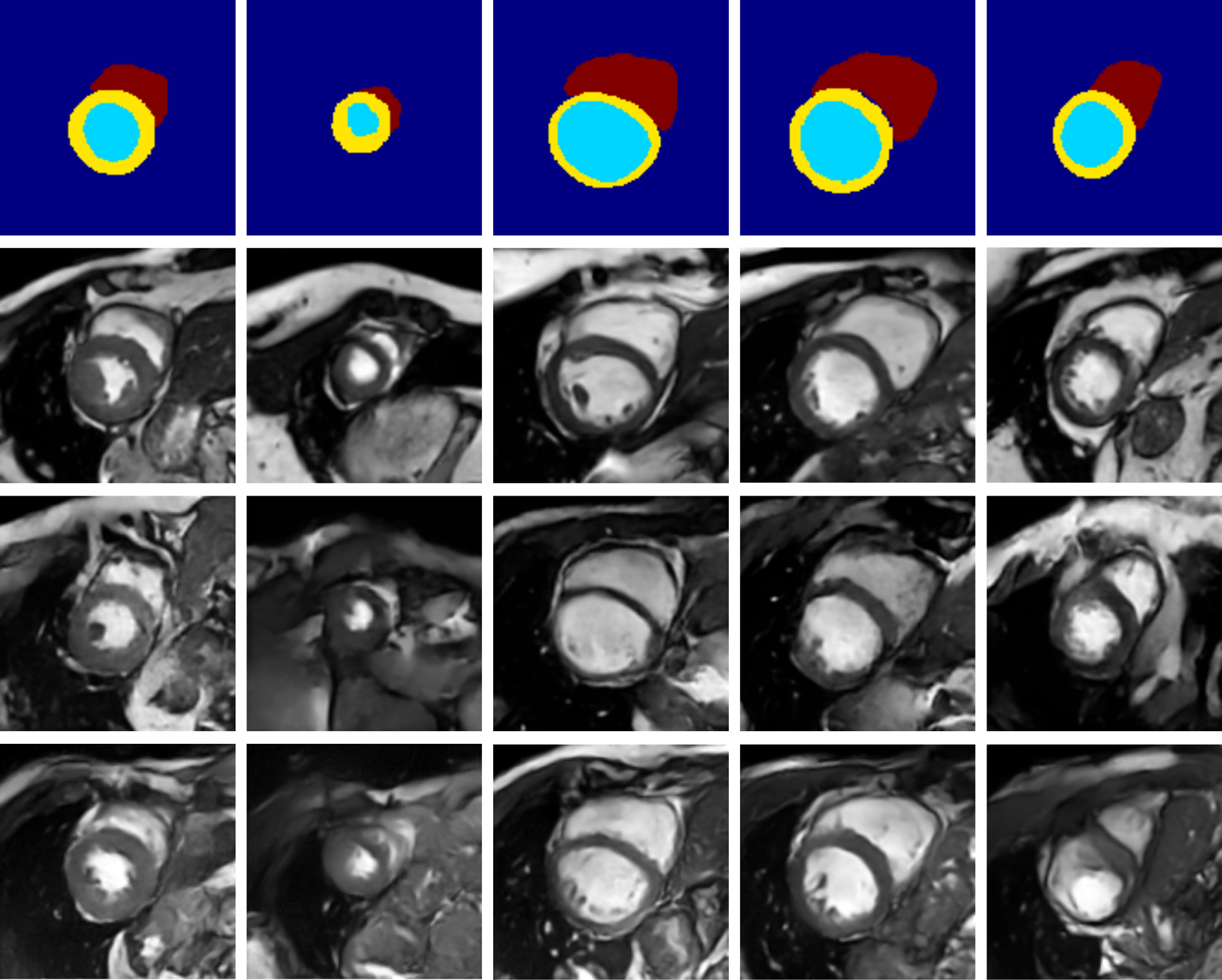}
    \caption{CMRI synthesis results.
Top row: input segmentation masks. Subsequent rows: synthetic images generated by DDPM, LDM, and FM respectively}
    \label{fig:results}
\end{figure}
\subsection{Anatomical \& Shape Plausibility of Generated Masks}
\begin{figure}[!htbp]
    \centering
    
    \begin{subfigure}{0.49\columnwidth}
        \centering
        \includegraphics[width=\linewidth]{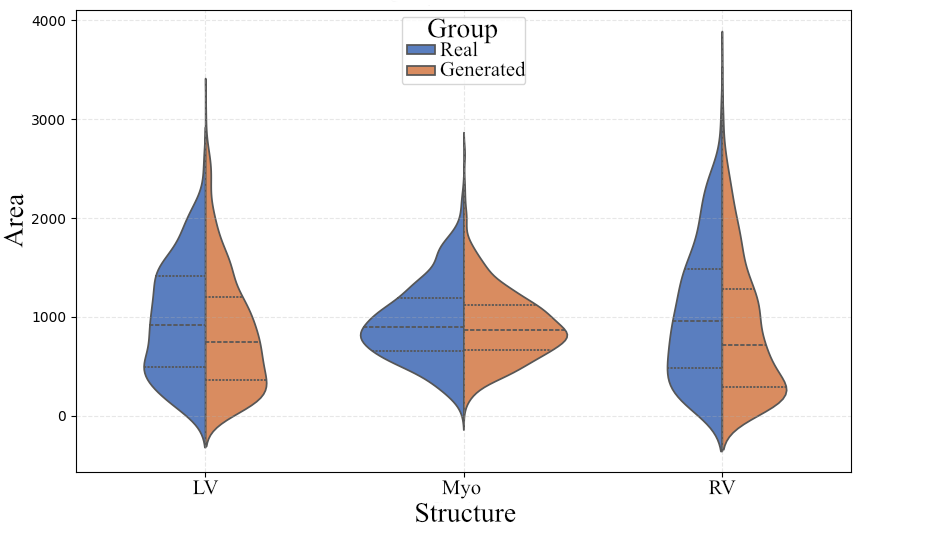}
        \caption{Area Distribution}
        \label{fig:violin_area}
    \end{subfigure}
    \hfill 
    \begin{subfigure}{0.49\columnwidth}
        \centering
        \includegraphics[width=\linewidth]{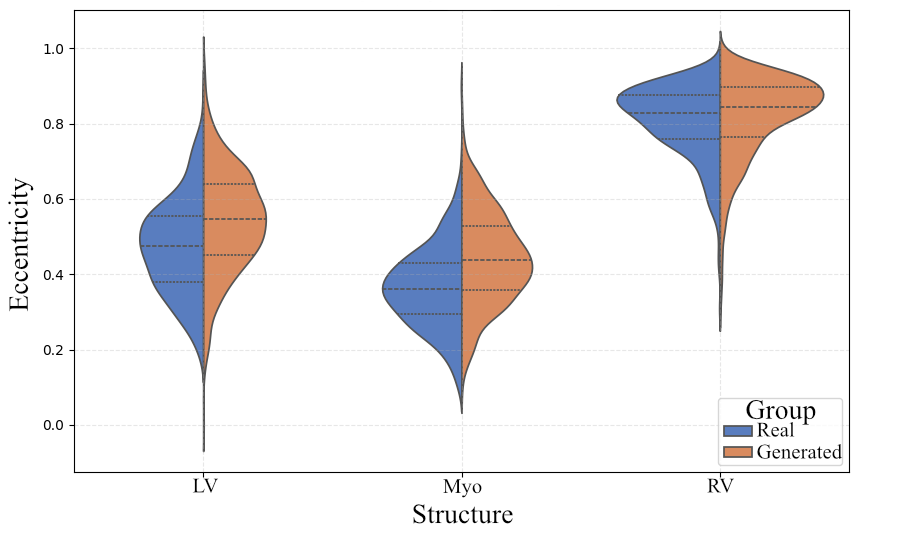}
        \caption{Eccentricity Distribution}
        \label{fig:violin_eccentricity}
    \end{subfigure}
    
    \vspace{0.3cm} 
    
    \begin{subfigure}{0.49\columnwidth}
        \centering
        \includegraphics[width=\linewidth]{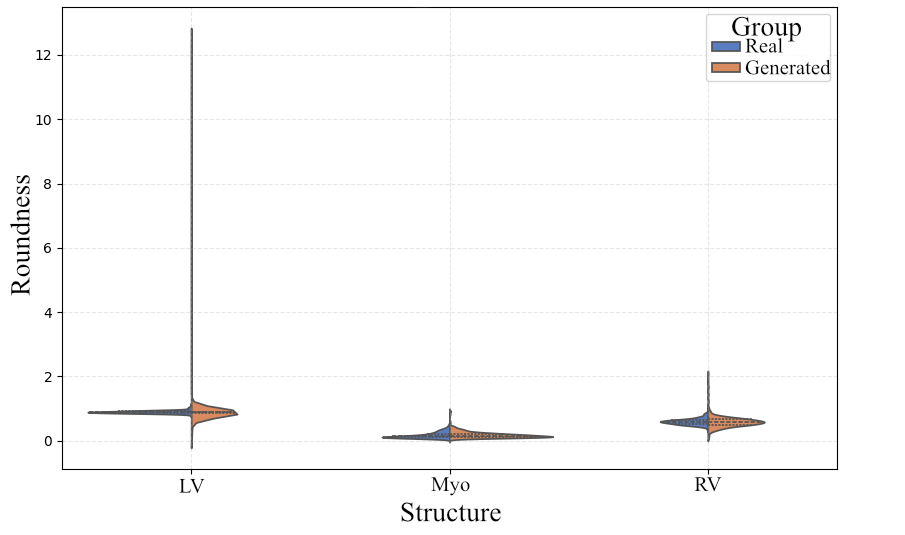}
        \caption{Roundness Distribution}
        \label{fig:violin_roundness}
    \end{subfigure}
    \hfill
    \begin{subfigure}{0.49\columnwidth}
        \centering
        \includegraphics[width=\linewidth]{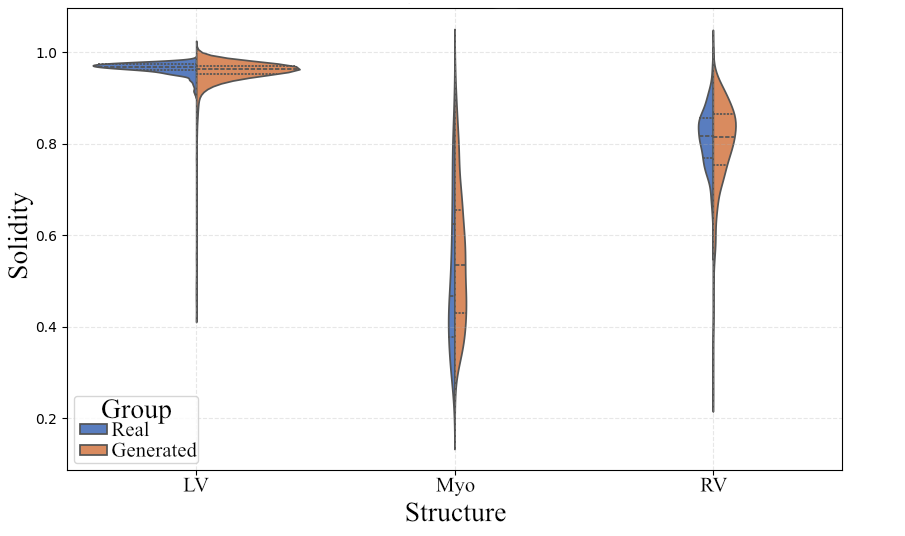}
        \caption{Solidity Distribution}
        \label{fig:violin_solidity}
    \end{subfigure}
    
    \caption{Comparison of geometric shape metrics between real and synthetic cardiac masks. (a) Area reflects the size of the segmented structure, (b) Eccentricity measures elongation, (c) Roundness evaluates circular compactness, and (d) Solidity assesses boundary smoothness and structural integrity. Overall, the real and synthetic distributions show strong alignment across most structures, with only minor deviations in (c) roundness and (d) solidity, indicating that the generated masks largely preserve anatomical geometry as assessed by (a)-(d).}
    \label{fig:shape_metrics_comparison}
\end{figure}

The distribution-based analysis shows that the generated masks largely preserve the anatomical characteristics of real cardiac structures (\textbf{Table}~\ref{tab:statistical_comparison}). While minor, statistically significant differences were observed in \textit{relative area} and certain shape descriptors (e.g., \textit{roundness} and \textit{solidity}), \textit{absolute area} and \textit{eccentricity} remained comparable in most cases (\textbf{Figure}~\ref{fig:shape_metrics_comparison}). Overall, the results indicate that the synthetic masks maintain realistic anatomical proportions with only moderate distributional variations.

\begin{table}[!htbp]
\caption{Statistical Comparison Between Real and Generated Masks}
\centering
\scriptsize
\begin{tabular}{|l|l|c|c|c|c|}
\hline
\textbf{Structure} & \textbf{Feature} & \textbf{Real Mean} & \textbf{Gen Mean} & \textbf{p-value}  \\
\hline
LV & Pixel\_Pct   & 5.1577 & 6.5380 & 0.0156 \\
LV & Area         & 845.04 & 875.31 & \textbf{0.9084}  \\
LV & Roundness    & 0.9009 & 0.7160 & $1.35\times10^{-40}$  \\
LV & Eccentricity & 0.4929 & 0.4728 & \textbf{0.1112} \\
LV & Solidity     & 0.9644 & 0.9001 & $1.35\times10^{-40}$ \\
\hline
RV & Pixel\_Pct   & 5.6487 & 5.9424 & 0.0061  \\
RV & Area         & 925.08 & 749.15 & \textbf{0.1548} \\
RV & Roundness    & 0.5997 & 0.5151 & $1.75\times10^{-6}$ \\
RV & Eccentricity & 0.8026 & 0.8196 & \textbf{0.3682} \\
RV & Solidity     & 0.8132 & 0.7727 & 0.0022  \\
\hline
Myo & Pixel\_Pct   & 5.8536 & 6.3840 & 0.0022  \\
Myo & Area         & 957.75 & 980.93 & \textbf{0.1113} \\
Myo & Roundness    & 0.1884 & 0.1311 & $8.00\times10^{-7}$ \\
Myo & Eccentricity & 0.3795 & 0.4017 & \textbf{0.0782} \\
Myo & Solidity     & 0.5366 & 0.5240 & \textbf{0.0539} \\
\hline
\end{tabular}
\label{tab:statistical_comparison}
\begin{tablenotes}
\footnotesize
\item ($p < 0.05$) indicates a statistically significant difference between real and synthetic distributions (Kolmogorov–Smirnov test). Non-significant values are highlighted in \textbf{bold}.
\end{tablenotes}
\end{table}

\subsection{Fidelity Evaluation of Generated Images}
Fidelity evaluation (\textbf{Table}~\ref{tab:fidelity_evaluation_results}) of 128x128 image generative models reveals distinct trade-offs among the three generative paradigms. DDPM achieves the strongest distribution-level realism, reflected in the lowest FID and KID, indicating close alignment with the real cardiac MRI data. In contrast, Flow Matching indicates improved perceptual and structural fidelity, achieving the highest MS-SSIM and lowest LPIPS, suggesting enhanced anatomical boundary preservation and texture realism. LDM has slightly reduced fidelity metrics, likely due to compression in latent space, although it offers better computational efficiency.

\begin{table}[!htbp]
\caption{Results of Fidelity Evaluation}
\centering
\scriptsize %
\renewcommand{\arraystretch}{1.1}
\setlength{\tabcolsep}{3pt}
\begin{tabularx}{\linewidth}{|p{1.4cm}|X|X|X|}

\hline
\textbf{Evaluation Metric} & \textbf{Diffusion-DDPM} & \textbf{Diffusion-LDM} & \textbf{Flow Match} \\
\hline
SSIM & \textbf{0.22} & 0.18 & \textbf{0.22} \\
\hline
MS-SSIM & 0.36 & 0.33 & \textbf{0.40} \\
\hline
PSNR & 10.67 & 9.95 & \textbf{11.44} \\
\hline
FID & \textbf{72.52} & 95.17 & 108.32 \\
\hline
KID & \textbf{0.04} & 0.08 & 0.098 \\
\hline
LPIPS & 0.49 & 0.51 & \textbf{0.48} \\
\hline
\end{tabularx}
\label{tab:fidelity_evaluation_results}
\end{table}

\subsection{Segmentation Utility Across Datasets}
Table~\ref{tab:utility_testing} shows that real-data training achieves the highest segmentation performance. In the full synthetic setting (synthetic masks and images), all generative models yield slightly lower Dice scores and higher boundary errors. DDPM, LDM, and Flow Matching achieve comparable overlap performance, though diffusion-based methods show slightly more stable boundary metrics. Dataset-specific synthetic training, where images are generated using real masks, consistently improves results and reduces the gap to real-data baselines. 

\begin{table}[!htbp]
\centering
\caption{Evaluation of Segmentation Model}
\label{tab:utility_testing}
\scriptsize
\renewcommand{\arraystretch}{1.1}
\setlength{\tabcolsep}{3pt}

\begin{tabular}{|p{2.6cm}|c|c|c|c|c|c|c|c|}
\hline

\multirow{2}{*}{\textbf{Training Setup}} 
& \multicolumn{8}{c|}{\textbf{Testing}} \\ \cline{2-9}

& \multicolumn{4}{c|}{\textbf{M\&M}} 
& \multicolumn{4}{c|}{\textbf{ACDC}} \\ \cline{2-9}

& Dice & IoU & HD95 & ASD
& Dice & IoU & HD95 & ASD \\ \hline

M\&M (Real)      
& 0.90 & 0.84 & 2.99  & 1.04
& 0.91 & 0.85 &2.89  &0.94  \\ \hline

ACDC (Real)      
& 0.90 & 0.83 &3.88  &1.23
& 0.93 & 0.88 &2.65  &0.75  \\ \hline

DDPM Full-Syn    
& 0.87 &0.80 &5.77  &1.79  
& 0.87 & 0.80 &6.28  &1.78  \\ \hline

DDPM ACDC-Syn     
&0.86  &0.79  &5.98  &1.89  
&0.88  &0.81  &4.72  &1.45  \\ \hline

DDPM M\&M-Syn      
&0.89  &\textbf{0.83}  &4.34  &1.41  
&\textbf{0.90}  &\textbf{0.84}  & 4.61 & \textbf{1.34} \\ \hline

LDM Full-Syn   
&0.87  &0.79  &5.28  &1.74  
&0.87  &0.80  &4.90  &1.52  \\ \hline

LDM ACDC-Syn    
&0.85  & 0.78 &7.94  & 2.48 
&0.86  &0.79  &8.38  &2.43  \\ \hline

LDM M\&M-Syn   
&\textbf{0.89}  &0.82  &\textbf{4.17}  & \textbf{1.40} 
& 0.88 &0.82  &\textbf{2.26}  & 1.53 \\ \hline

FM Full-Syn   
&0.87  &0.80  &6.53  &2.12  
&0.88  &0.81  & 6.29 & 1.81 \\ \hline

FM ACDC-Syn   
&0.82  &0.73  &8.75  & 2.90 
&0.85  &0.76  &7.74  & 2.19 \\ \hline

FM M\&M-Syn  
&0.88  &0.82  &5.04  & 1.64 
& 0.89 &0.82  &5.73  &1.67  \\ \hline

\end{tabular}
\begin{tablenotes}
\footnotesize
\item \textit{Full-Syn}: Generated images conditioned synthetic masks only; 
\textit{ACDC-Syn}: Generated images conditioned ACDC training masks; 
\textit{M\&M-Syn}: Generated images conditioned M\&M training masks. Best results among synthetic training setups are shown in \textbf{bold}.
\end{tablenotes}
\end{table}

\subsection{Privacy Concerns}
Privacy evaluation (\textbf{Table}~\ref{tab:privacy_evaluation_results}) indicates that all three generative models preserve patient privacy effectively. Nearest Neighbor analysis (L2, LPIPS, NNDR) confirms that the models do not memorize or replicate specific training samples, generating anatomically plausible but perceptually distinct images. Membership Inference Attack (MIA) yields AUC scores near 0.5 (specifically 0.58-0.60), indicating that the models are robust against re-identification attacks. Among the models, LDM offers the strongest privacy guarantee with lowest AUC value, followed by FM and DDPM with slightly higher but still safe values~\cite{Hosmer2013}.

\begin{table}[!htbp]
\caption{Results of Privacy Evaluation. ROC-AUC values are from Membership Inference Attack. }
\centering
\scriptsize %
\renewcommand{\arraystretch}{1.1}
\setlength{\tabcolsep}{3pt}
\begin{tabularx}{\linewidth}{|p{2cm}|X|X|X|}

\hline
\textbf{Evaluation Metric} & \textbf{Diffusion-DDPM} & \textbf{Diffusion-LDM} & \textbf{Flow Match} \\
\hline
Nearest Neighbor (L2 Distance) & 12.0 & 10.5 & \textbf{19.0} \\
\hline
LPIPS & 0.36 & 0.37 & \textbf{0.41} \\
\hline
NNDR & 0.83 & 0.85 & \textbf{0.87} \\
\hline
ROC\_AUC (MIA)  & 0.6029 & \textbf{0.580} & 0.6038 \\
\hline
\end{tabularx}
\label{tab:privacy_evaluation_results}
\end{table}

\section{Conclusion}



Results across fidelity, segmentation utility, and privacy evaluation indicate that the proposed generative framework achieves consistently competitive performance in all three defined assessment dimensions. The generated data preserve anatomical realism, support effective downstream segmentation, and demonstrate robustness against privacy leakage. Overall, these findings suggest that the proposed approach can serve as a reliable and valid strategy for synthetic data augmentation in CMR.

\section*{Acknowledgment}
This work is part of the European project SEARCH, which is supported by the Innovative Health Initiative Joint Undertaking (IHI JU) under grant agreement No. 101172997. The JU receives support from the European Union’s Horizon Europe research and innovation programme and COCIR, EFPIA, Europa Bio, MedTech Europe, Vaccines Europe, Medical Values GmbH, Corsano Health BV, Syntheticus AG, Maggioli SpA, Motilent Ltd, Ubitech Ltd, Hemex Benelux, Hellenic Healthcare Group, German Oncology Center, Byte Solutions Unlimited, AdaptIT GmbH. Views and opinions expressed are however those of the author(s) only and do not necessarily reflect those of the aforementioned parties. Neither of the aforementioned parties can be held responsible for them. 


\bibliographystyle{IEEEtran}
\bibliography{export}

\end{document}